\documentclass{article}
\usepackage{kr}

\usepackage{times}
\usepackage{soul}
\usepackage{url}
\usepackage[hidelinks]{hyperref}
\usepackage[utf8]{inputenc}
\usepackage[small]{caption}
\usepackage{graphicx}
\usepackage{amsmath}
\usepackage{amsthm}
\usepackage{booktabs}
\usepackage{algorithm}
\usepackage{algorithmic}

\usepackage[table]{xcolor}
\usepackage{xcolor}
\usepackage{multirow}
\usepackage{subcaption}
\usepackage{caption}
\usepackage{colortbl}
\usepackage{flafter}
\usepackage{amssymb}
\usepackage{stfloats}
\usepackage[export]{adjustbox}

\title{SynergyKGC: Reconciling Topological Heterogeneity in Knowledge Graph Completion via Topology-Aware Synergy}

\author{%
    Xuecheng Zou$^1$\and
    Yu Tang$^{1,}$\thanks{Corresponding author.}\and
    Bingbing Wang$^2$ \\
    \affiliations
    $^1$School of Future Science and Engineering, Soochow University\\
    $^2$School of Mathematical Sciences, Soochow University\\
    \emails
    \{xczouxczou, bbwangstat1\}@stu.suda.edu.cn,
    ytang@suda.edu.cn
}

\begin{document}

\maketitle

\begin{abstract}
Knowledge Graph Completion (KGC) fundamentally hinges on the coherent fusion of pre-trained entity semantics with heterogeneous topological structures to facilitate robust relational reasoning. However, existing paradigms encounter a critical "structural resolution mismatch," failing to reconcile divergent representational demands across varying graph densities, which precipitates structural noise interference in dense clusters and catastrophic representation collapse in sparse regions. We present \textit{SynergyKGC}, an adaptive framework that advances traditional neighbor aggregation to an active Cross-Modal Synergy Expert via relation-aware cross-attention and semantic-intent-driven gating. By coupling a density-dependent Identity Anchoring strategy with a Double-tower Coherent Consistency architecture, SynergyKGC effectively reconciles topological heterogeneity while ensuring representational stability across training and inference phases. Systematic evaluations on two public benchmarks validate the superiority of our method in significantly boosting KGC hit rates, providing empirical evidence for a generalized principle of resilient information integration in non-homogeneous structured data.
\end{abstract}

\section{Introduction}

Knowledge Graph Completion (KGC) has emerged as a pivotal task for expanding large-scale relational networks and supporting downstream reasoning~\cite{hogan:csur21-kg,yasunaga:naacl21-qagnn,guo:tkde22-survey,dong:kdd14-knowledgevault}, which has evolved from a relational modeling task into a complex multi-modal reconciliation challenge, necessitating the seamless alignment of textual semantics and topological structures. While current paradigms leverage Pre-trained Language Models (PLMs) and Graph Neural Networks (GNNs) to capture these disparate signals, a fundamental bottleneck remains: Passive Structural-Semantic Alignment. Existing hybrid models typically treat structural contexts as static features, failing to adapt to the inherent topological heterogeneity of different knowledge domains. This rigidity leads to two critical pathologies: (i) Representation Collapse in sparse hierarchies where entities lack sufficient structural scaffolding, and (ii) Identity Redundancy in dense clusters where explicit structural cues introduce irreducible noise.

To address these challenges, we propose \textit{SynergyKGC}, a high-precision framework designed for Instruction-Driven Structural-Semantic Synergy. A defining hallmark of our framework is the ``catch-up effect'' triggered instantaneously upon the activation of the Synergy Expert. Unlike traditional methods that suffer from prohibitive computational overhead due to exhaustive warming phases (often exceeding 30 epochs), \textit{SynergyKGC} utilizes semantic intent as an active query instruction to retrieve and filter structural cues. This mechanism abruptly synchronizes the disparate semantic and structural streams—which diverge during initial training—into a coherent, high-fidelity manifold, significantly accelerating both convergence and prediction precision.

The architectural integrity of \textit{SynergyKGC} is governed by the principle of Dual-Axis Consistency. On the \textit{architectural axis}, we enforce strict synergy alignment between the Query and Entity towers to ensure symmetric representation learning. On the \textit{lifecycle axis}, we maintain consistency across the training and inference phases, effectively mitigating the \textbf{representation drift} that typically degrades performance during deployment. Furthermore, we introduce the Identity Anchoring (IA) strategy and its underlying theory, defining the \textit{Structure $\approx$ Identity} phenomenon: the insight that while explicit identity signals are redundant in dense graphs, they serve as essential positional scaffolding in sparse environments.

Experimental results on FB15k-237 and WN18RR rigorously validate our approach, with \textit{SynergyKGC} establishing a new state-of-the-art through a remarkable +8.0\% absolute gain in Hits@1 on WN18RR. This breakthrough is underpinned by three synergistic pillars: first, the \textit{SynergyKGC} framework itself, an instruction-driven paradigm that adaptively reconciles dual-modal signals via active structural retrieval; second, the principle of Dual-Axis Consistency, which enforces rigorous alignment across both the dual-tower architecture and the training-inference lifecycle to eliminate the pervasive distribution shift; and furthermore, the theory of Identity Anchoring, which formalizes the density-driven trade-off between structural sufficiency and identity redundancy to resolve the topological heterogeneity inherent in knowledge graph completion. Finally, superior computational efficiency, manifested by a unique ``catch-up effect'' that enables the model to bypass exhaustive warming phases and achieve rapid convergence, significantly reducing training overhead while enhancing predictive precision.

\begin{figure*}[!t]
  \centering
  \includegraphics[width=0.88\textwidth]{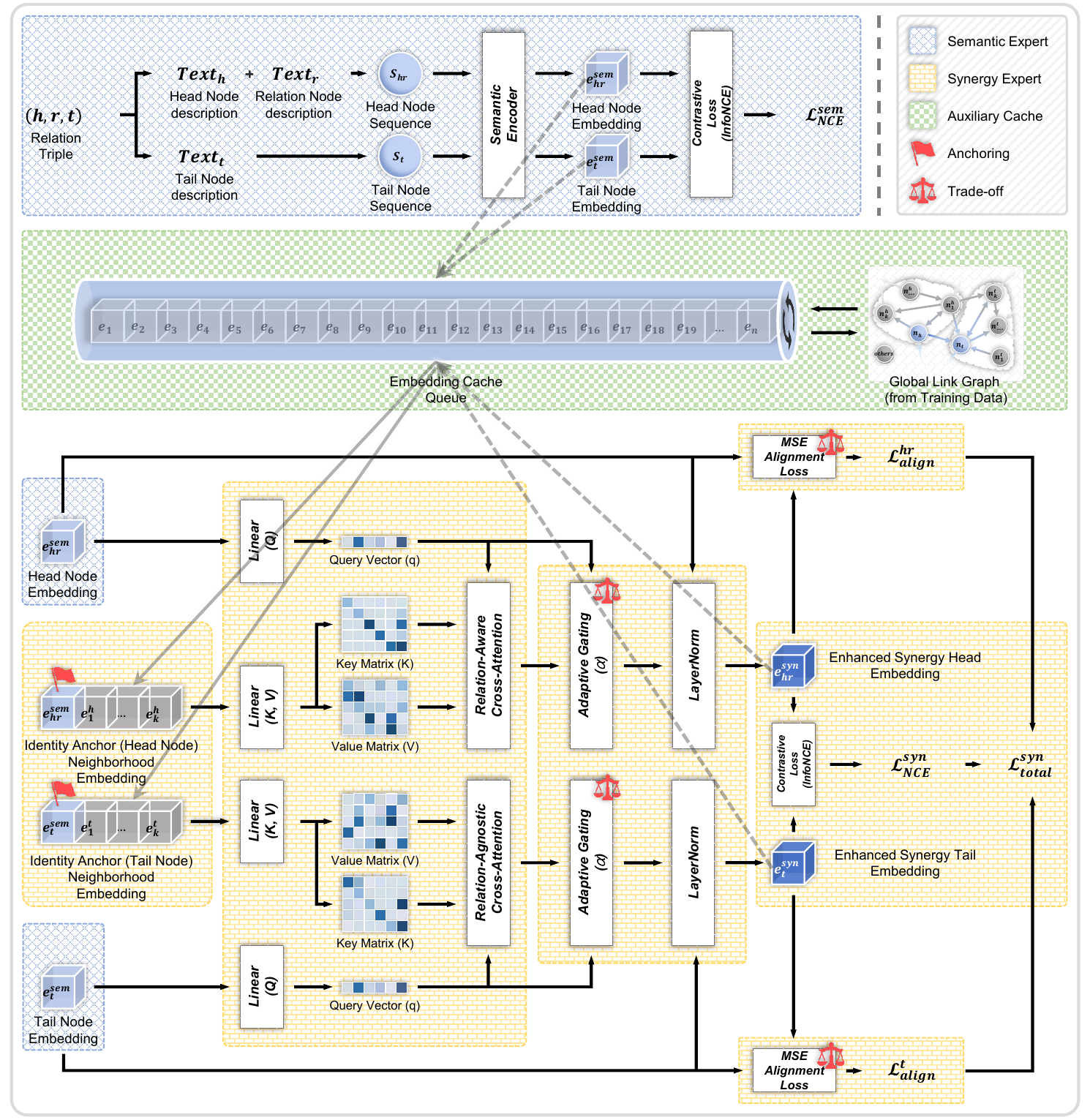}
  \caption{The overall architecture of SynergyKGC. The framework establishes a progressive cross-modal reconciliation between latent semantic manifolds and explicit topological signals, where initial textual embeddings ($e^{sem}$) from the Semantic Expert are enhanced by the Synergy Expert via an asymmetric dual-tower design: the $(h,r)$ stream employs \textit{Relation-Aware Cross-Attention} for context retrieval, while the $t$ stream utilizes a \textit{Relation-Agnostic} counterpart to maintain representation consistency during inference. This integration is governed by a density-adaptive identity anchoring strategy that dynamically toggles entity-self signals based on a dataset-specific threshold $\phi$ ($\phi=1$ for dense FB15k-237; no threshold for sparse WN18RR) to prevent representation collapse while suppressing structural noise, all within a joint optimization framework where adaptive gating ($\alpha$) is constrained by contrastive ($\mathcal{L}_{NCE}$) and MSE alignment ($\mathcal{L}_{align}$) objectives to effectively mitigate semantic drift during structural-semantic fusion.}
  \label{fig:framework}
\end{figure*}

\section{Related Work}

\subsection{Embedding-based KGC}
Early KGC paradigms primarily focused on learning low-dimensional latent representations through structural translation (e.g., TransE~\cite{bordes:nips13-transe,wang:aaai14-transh,lin:aaai15-transr}) or rotational transformations (e.g., RotatE~\cite{sun:iclr19-rotate}). Bilinear models such as DistMult~\cite{yang:iclr15-distmult} and TuckER~\cite{balazevic:emnlp19-tucker} further expanded this by capturing higher-order relational interactions. While effective in dense graphs (e.g., FB15k-237), these methods are inherently limited by their lack of semantic understanding and often suffer from representation collapse in sparse environments like WN18RR, where structural patterns are insufficient.

\begin{table}[!t]
  \centering
  \scriptsize
  \definecolor{colorBlue}{HTML}{446A91}
  \definecolor{colorGreen}{HTML}{6DC170}
  \definecolor{colorYellow}{HTML}{E6B400}
  \definecolor{colorRed}{HTML}{FF4D4F}
  \definecolor{colorGray}{HTML}{4B5563}
  \newcommand{\deep}{38}
  \newcommand{\light}{10}

  \renewcommand{\arraystretch}{1.2}
  \setlength{\tabcolsep}{10pt}

  \begin{tabular}{llcc}
    \toprule
    & & \multicolumn{2}{c}{\textbf{Datasets}} \\
    \cmidrule(lr){3-4}
    \textbf{Category} & \textbf{Metric} & \textbf{FB15k-237} & \textbf{WN18RR} \\
    \midrule

    \multirow{3}{*}{\shortstack[l]{\textbf{General}\\\textit{\textcolor{colorGray}{Topology}}}}
    & \# Entities & \cellcolor{colorBlue!\light}14,541 & \cellcolor{colorBlue!\deep}40,943 \\
    & \# Relations & \cellcolor{colorGreen!\deep}237 & \cellcolor{colorGreen!\light}11 \\
    & \# Train & \cellcolor{colorYellow!\deep}272,115 & \cellcolor{colorYellow!\light}86,835 \\
    & \# Valid & \cellcolor{colorRed!\deep}17,535 & \cellcolor{colorRed!\light}3,034 \\
    & \# Test & \cellcolor{colorGray!\deep}20,466 & \cellcolor{colorGray!\light}3,134 \\

    \midrule

    \multirow{3}{*}{\shortstack[l]{\textbf{Textual}\\\textit{\textcolor{colorGray}{Semantic}}}}
    & Avg. Token
Counts & \cellcolor{colorBlue!\deep}114.6 & \cellcolor{colorBlue!\light}17.1 \\
    & Med. Token
Counts & \cellcolor{colorGreen!\deep}76 & \cellcolor{colorGreen!\light}15 \\
    & Vocab Size & \cellcolor{colorYellow!\deep}28,538 & \cellcolor{colorYellow!\light}18,741 \\

    \midrule

    \multirow{3}{*}{\shortstack[l]{\textbf{Density}\\\textit{\textcolor{colorGray}{Percentile}}}}
    & $P_{1} / P_{10}$ & \cellcolor{colorBlue!\deep}1 / 5 & \cellcolor{colorBlue!\light}1 / 1 \\
    & $P_{25} / P_{50}$ & \cellcolor{colorGreen!\deep}11 / 22 & \cellcolor{colorGreen!\light}2 / 3 \\
    & $P_{75} / P_{90}$ & \cellcolor{colorYellow!\deep}41 / 68 & \cellcolor{colorYellow!\light}5 / 8 \\
    & Max ($P_{100}$) & \cellcolor{colorRed!\deep}7,614 & \cellcolor{colorRed!\light}482 \\

    \bottomrule
  \end{tabular}

  \vspace{4pt}
  \caption{Statistical profiles of FB15k-237 and WN18RR. Metrics characterize topological, semantic, and density attributes. Shading contrasts the relational density and textual verbosity of FB15k-237 against the structural sparsity and semantic brevity of WN18RR; $P_x$ denotes entity degree percentiles.}
  \label{tab:dataset_stats_compact}
\end{table}

\begin{table*}[!t]
  \centering
  \scriptsize

  \definecolor{colorBlue}{HTML}{446A91}
  \definecolor{colorGreen}{HTML}{6DC170}
  \definecolor{colorYellow}{HTML}{E6B400}
  \definecolor{colorRed}{HTML}{FF4D4F}
  \definecolor{colorGray}{HTML}{4B5563}
  \definecolor{incrc}{HTML}{B91C1C}
  \newcommand{\deep}{38}
  \newcommand{\light}{10}

  \setlength{\tabcolsep}{8pt}
  \begin{tabular}{@{} c c @{}}

    \begin{minipage}[t]{0.58\textwidth}
      \vspace{0pt}
      \centering
      \renewcommand{\arraystretch}{1.2}
      \setlength{\tabcolsep}{5.0pt}

      \begin{tabular}{l cccc cccc}
        \toprule
        \textbf{Method} & \multicolumn{4}{c}{\textbf{FB15k-237}} & \multicolumn{4}{c}{\textbf{WN18RR}} \\
        \cmidrule(lr){2-5} \cmidrule(lr){6-9}
        & \textbf{MRR} & \textbf{H@1} & \textbf{H@3} & \textbf{H@10} & \textbf{MRR} & \textbf{H@1} & \textbf{H@3} & \textbf{H@10} \\
        \midrule

        \textit{\textcolor{colorGray}{Embedding-based}} \\
        TransE~\shortcite{bordes:nips13-transe}   & \cellcolor{colorBlue!\light}27.9 & \cellcolor{colorGreen!\light}19.8 & \cellcolor{colorYellow!\light}37.6 & \cellcolor{colorRed!\light}44.1 & \cellcolor{colorBlue!\light}24.3 & \cellcolor{colorGreen!\light}4.3  & \cellcolor{colorYellow!\light}44.1 & \cellcolor{colorRed!\light}53.2 \\
        RotatE~\shortcite{sun:iclr19-rotate}   & \cellcolor{colorBlue!\light}47.6 & \cellcolor{colorGreen!\light}42.8 & \cellcolor{colorYellow!\light}49.2 & \cellcolor{colorRed!\light}57.1 & \cellcolor{colorBlue!\light}33.8 & \cellcolor{colorGreen!\light}24.1 & \cellcolor{colorYellow!\light}37.5 & \cellcolor{colorRed!\light}53.3 \\
        DistMult~\shortcite{yang:iclr15-distmult} & \cellcolor{colorBlue!\light}28.1 & \cellcolor{colorGreen!\light}19.9 & \cellcolor{colorYellow!\light}30.1 & \cellcolor{colorRed!\light}44.6 & \cellcolor{colorBlue!\light}44.4 & \cellcolor{colorGreen!\light}41.2 & \cellcolor{colorYellow!\light}47.0 & \cellcolor{colorRed!\light}50.4 \\
        ComplEx~\shortcite{trouillon:icml16-complex}  & \cellcolor{colorBlue!\light}27.8 & \cellcolor{colorGreen!\light}19.4 & \cellcolor{colorYellow!\light}29.7 & \cellcolor{colorRed!\light}45.0 & \cellcolor{colorBlue!\light}44.9 & \cellcolor{colorGreen!\light}40.9 & \cellcolor{colorYellow!\light}46.9 & \cellcolor{colorRed!\light}53.0 \\
        TuckER~\shortcite{balazevic:emnlp19-tucker}   & \cellcolor{colorBlue!\light}35.8 & \cellcolor{colorGreen!\light}26.6 & \cellcolor{colorYellow!\light}39.4 & \cellcolor{colorRed!\light}54.4 & \cellcolor{colorBlue!\light}47.0 & \cellcolor{colorGreen!\light}44.3 & \cellcolor{colorYellow!\light}48.2 & \cellcolor{colorRed!\light}52.6 \\
        \midrule

        \textit{\textcolor{colorGray}{Text-based / Hybrid}} \\
        KG-BERT~\shortcite{yao:arxiv19-kgbert}  & \cellcolor{colorBlue!\light}--   & \cellcolor{colorGreen!\light}--   & \cellcolor{colorYellow!\light}--   & \cellcolor{colorRed!\light}42.0 & \cellcolor{colorBlue!\light}21.6 & \cellcolor{colorGreen!\light}4.1  & \cellcolor{colorYellow!\light}30.2 & \cellcolor{colorRed!\light}52.4 \\
        StAR~\shortcite{wang:www21-star}     & \cellcolor{colorBlue!\light}29.6 & \cellcolor{colorGreen!\light}20.5 & \cellcolor{colorYellow!\light}32.2 & \cellcolor{colorRed!\light}48.2 & \cellcolor{colorBlue!\light}40.1 & \cellcolor{colorGreen!\light}24.3 & \cellcolor{colorYellow!\light}49.1 & \cellcolor{colorRed!\light}70.9 \\
        LP-BERT~\shortcite{li:tallip23-lpbert}  & \cellcolor{colorBlue!\light}31.0 & \cellcolor{colorGreen!\light}22.3 & \cellcolor{colorYellow!\light}33.6 & \cellcolor{colorRed!\light}49.0 & \cellcolor{colorBlue!\light}48.2 & \cellcolor{colorGreen!\light}34.3 & \cellcolor{colorYellow!\light}56.3 & \cellcolor{colorRed!\light}75.2 \\
        KG-S2S~\shortcite{chen:coling22-kgs2s}   & \cellcolor{colorBlue!\light}33.6 & \cellcolor{colorGreen!\light}25.6 & \cellcolor{colorYellow!\light}37.1 & \cellcolor{colorRed!\light}49.8 & \cellcolor{colorBlue!\light}57.4 & \cellcolor{colorGreen!\light}53.1 & \cellcolor{colorYellow!\light}59.5 & \cellcolor{colorRed!\light}66.1 \\
        SimKGC~\shortcite{wang:acl22-simkgc}   & \cellcolor{colorBlue!\light}33.6 & \cellcolor{colorGreen!\light}24.9 & \cellcolor{colorYellow!\light}36.2 & \cellcolor{colorRed!\light}51.1 & \cellcolor{colorBlue!\light}66.6 & \cellcolor{colorGreen!\light}58.7 & \cellcolor{colorYellow!\light}71.7 & \cellcolor{colorRed!\light}80.5 \\
        ProgKGC~\shortcite{li:iswc25-progkgc}  & \cellcolor{colorBlue!\light}34.4 & \cellcolor{colorGreen!\light}25.5 & \cellcolor{colorYellow!\light}37.5 & \cellcolor{colorRed!\light}52.3 & \cellcolor{colorBlue!\light}68.2 & \cellcolor{colorGreen!\light}59.7 & \cellcolor{colorYellow!\light}73.9 & \cellcolor{colorRed!\light}83.4 \\
        \midrule

        \textbf{SynergyKGC (Ours)}
        & \cellcolor{colorBlue!\deep}\textbf{39.9}
        & \cellcolor{colorGreen!\deep}\textbf{30.2}
        & \cellcolor{colorYellow!\deep}\textbf{43.6}
        & \cellcolor{colorRed!\deep}\textbf{59.4}
        & \cellcolor{colorBlue!\deep}\textbf{74.2}
        & \cellcolor{colorGreen!\deep}\textbf{67.7}
        & \cellcolor{colorYellow!\deep}\textbf{78.5}
        & \cellcolor{colorRed!\deep}\textbf{85.5} \\

        \textit{Improvement}
        & \textbf{\textcolor{incrc}{\textit{+5.5}}}
        & \textbf{\textcolor{incrc}{\textit{+4.7}}}
        & \textbf{\textcolor{incrc}{\textit{+6.1}}}
        & \textbf{\textcolor{incrc}{\textit{+7.1}}}
        & \textbf{\textcolor{incrc}{\textit{+6.0}}}
        & \textbf{\textcolor{incrc}{\textit{+8.0}}}
        & \textbf{\textcolor{incrc}{\textit{+4.6}}}
        & \textbf{\textcolor{incrc}{\textit{+2.1}}} \\
        \bottomrule
      \end{tabular}
    \end{minipage}
    &
    \begin{minipage}[t]{0.38\textwidth}
      \vspace{0pt}
      \centering
      \includegraphics[width=\linewidth]{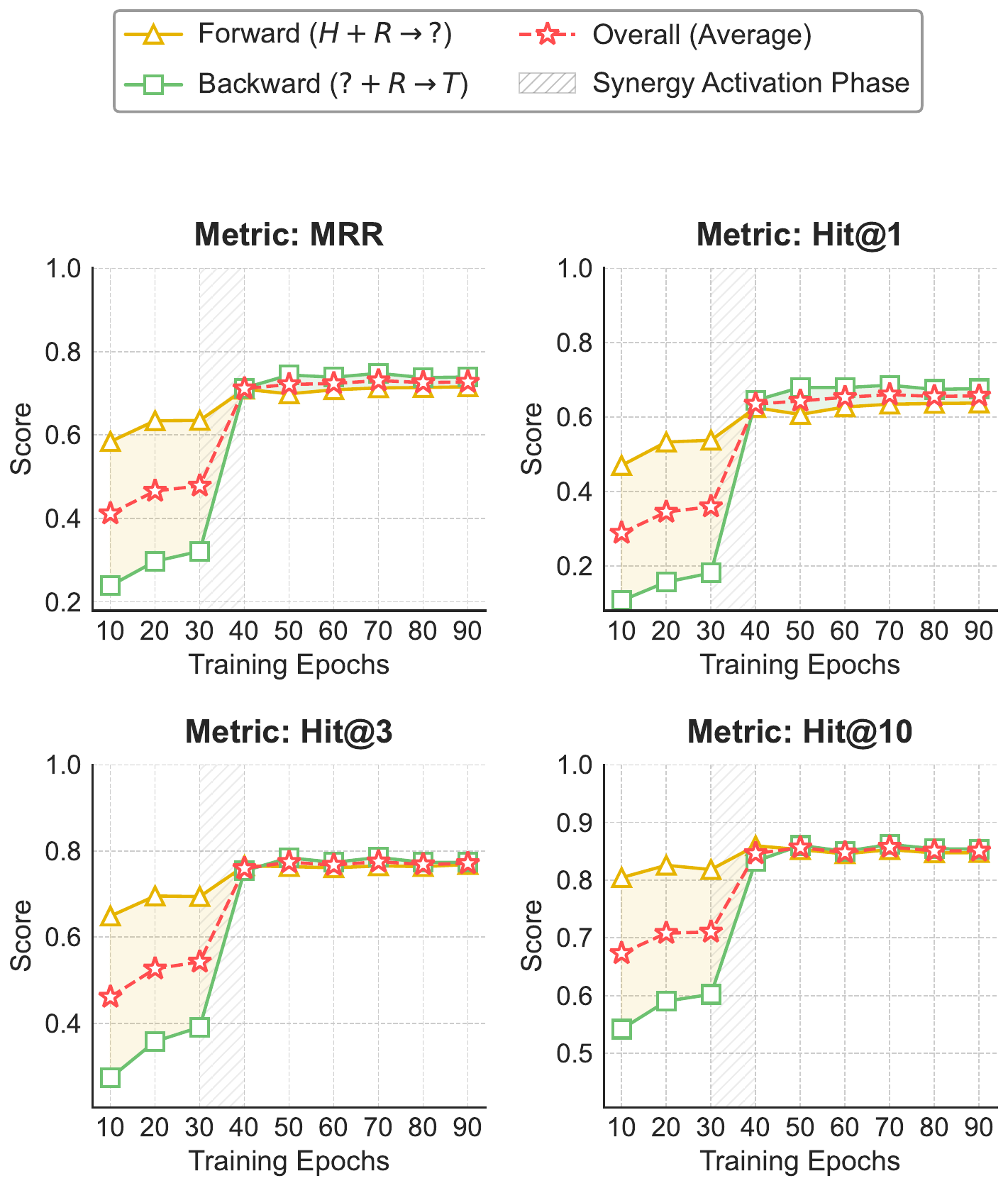}
    \end{minipage}
    \\

    \begin{minipage}[b]{0.58\textwidth}
        \captionof{table}{Main link prediction results on FB15k-237 and WN18RR. Bold values denote the best performance; \textit{Improvement} is calculated against the strongest hybrid baseline. \textit{SynergyKGC} achieves a comprehensive breakthrough, significantly outperforming state-of-the-art text-based models. Notably, the substantial gains in precision, highlighted by a +8.0\% Hits@1 improvement on WN18RR, validate the model's superior capacity to reconcile semantic and structural modalities across varying graph densities.}
        \label{tab:main_results}
    \end{minipage}
    &
    \begin{minipage}[b]{0.38\textwidth}
        \captionof{figure}{Instantaneous bidirectional synchronization triggered by the Synergy Expert (WN18RR). The activation at Epoch 30 (shaded) abruptly eliminates the representation gap, driving the lagging backward stream to synchronize with the forward stream. This ``catch-up'' effect demonstrates that late-phase synergy effectively reconciles dual-tower representations into a coherent, high-precision state.}
        \label{fig:mrr_comparison}
    \end{minipage}

  \end{tabular}
\end{table*}

\subsection{PLM-based and Hybrid KGC}
The advent of Pre-trained Language Models (PLMs) has shifted the research focus toward leveraging textual descriptions. SimKGC~\cite{wang:acl22-simkgc,wang:ijcai19-generative,he:icme24-mocosa} utilized contrastive learning to establish a strong semantic baseline, while subsequent works like ProgKGC~\cite{wang:wsdm22-nafet,borrego:eaai21-cafe,velickovic:iclr18-gat,li:iswc25-progkgc} attempted to incorporate neighborhood information via phased training. However, these hybrid approaches typically perform \textit{passive structural fusion}, where topological signals are treated as static add-ons. This design not only triggers Manifold Drift during the transition from semantic warming to structural integration but also necessitates an exhaustive warming phase (often exceeding 30 epochs) to stabilize the base encoder, leading to prohibitive computational overhead and suboptimal training efficiency.

\subsection{Gaps in Synergy and Consistency}
A critical yet overlooked challenge in current KGC research is the Inference-time Distribution Shift. Most neighborhood-aware models are hindered by two structural flaws: (i) \textit{Relational Asymmetry}, where they predominantly focus on head-entity neighborhoods while neglecting the crucial topological context of tail candidates; and (ii) \textit{Structural Decoupling}, where they disable structural aggregation during inference to save costs, leading to a significant representation gap between training and deployment.

In contrast to these paradigms, \textit{SynergyKGC} establishes the principle of \textbf{Dual-Axis Consistency}. By ensuring strict synergy alignment across both the dual-tower architecture and the training-inference lifecycle, we eliminate manifold drift and the need for exhaustive warming phases. Specifically, by replacing passive aggregation with \textbf{instruction-driven retrieval} and incorporating \textbf{density-aware identity anchoring}, our framework triggers an instantaneous \textit{catch-up effect} that synchronizes disparate modal streams efficiently. This holistic approach effectively resolves the trade-off between structural sufficiency and identity redundancy, delivering paradigm-shifting performance, notably a \textbf{+8.0\% absolute gain in Hits@1} on sparse benchmarks like WN18RR.

\section{Methodology}
\label{sec:method}

We propose \textit{SynergyKGC}, a progressive dual-tower framework designed to reconcile latent semantic manifolds with explicit topological signals, as illustrated in Figure~\ref{fig:framework}. Built upon a two-phase training paradigm, the architecture systematically integrates a Semantic Expert for foundational linguistic priors with a bidirectional Synergy Expert that adaptively fuses structural contexts via relation-aware cross-attention and density-aware identity anchoring. To ensure representation stability, we employ semantic consistency regularization (via MSE alignment loss) alongside dynamic dual-tower consistency to eliminate inference-time distribution discrepancies and mitigate manifold drift during structural-semantic integration.

\noindent \textbf{Problem Formulation.} Formally, let $\mathcal{G}=(\mathcal{E}, \mathcal{R}, \mathcal{T})$ denote a Knowledge Graph (KG), where $\mathcal{E}$ and $\mathcal{R}$ represent the sets of entities and relations, respectively, and $\mathcal{T} \subseteq \mathcal{E} \times \mathcal{R} \times \mathcal{E}$ is the set of observed triplets. Given an incomplete triplet $(h, r, ?)$, the objective of link prediction is to learn a scoring function $\psi(h, r, t)$ that ranks the ground-truth tail entity $t \in \mathcal{E}$ above all other candidates $t' \in \mathcal{E} \setminus \{t\}$~\cite{ji:tnnls22-survey}.

\subsection{Semantic Expert}

In Phase I, we initialize the Semantic Expert to establish a high-fidelity \textit{latent semantic manifold}, providing a robust foundational reference for subsequent structural synergy. This pre-conditioning ensures that the representation space is sufficiently aligned before the introduction of topological signals, thereby mitigating potential manifold drift.

\noindent \textbf{Textual Encoding.} We leverage a pre-trained BERT~\cite{devlin:naacl19-bert} as the backbone dual-tower encoder. To capture relation-aware semantic intent, we map the triplet text into low-dimensional latent vectors via the following encoding scheme:
\begin{equation}
    S_{hr} = [\text{CLS}] \oplus \text{Text}_h \oplus [\text{SEP}] \oplus \text{Text}_r,
\end{equation}
\begin{equation}
    S_{t} = [\text{CLS}] \oplus \text{Text}_t,
\end{equation}
\begin{equation}
    \mathbf{e}_{x}^{sem} = \text{BERT}_{\theta}\left( S_{x} \right), \quad x \in \{hr, t\},
\end{equation}
where $\oplus$ denotes sequence concatenation, and $\mathbf{e}_{x}^{sem} \in \mathbb{R}^d$ represents the initial semantic embedding optimized for linguistic discriminability.

\noindent \textbf{Warming Objective.} We define the compatibility score in the semantic space using cosine similarity $s_{sem}$:
\begin{equation}
    s_{sem}(h,r,t) = \frac{(\mathbf{e}_{hr}^{sem})^\top \mathbf{e}_{t}^{sem}}{\| \mathbf{e}_{hr}^{sem} \|_2 \| \mathbf{e}_{t}^{sem} \|_2}.
\end{equation}
Subsequently, the model is warmed up via the contrastive loss (InfoNCE) to achieve a discriminative spatial distribution across the batch $\mathcal{B}$:
\begin{equation}
    \mathcal{L}_{NCE}^{sem} = - \sum_{(h,r,t) \in \mathcal{B}} \log \frac{e^{s_{sem}(h,r,t) / \tau}}{\sum_{t' \in \mathcal{B}_{neg} \cup \{t\}} e^{s_{sem}(h,r,t') / \tau}}
\end{equation}
where $\tau$ is the temperature hyperparameter. This warming phase yields a stable semantic distribution as a prerequisite for executing instruction-driven structural retrieval, effectively preventing the model from being distracted by topological noise during the nascent stages of synergy.

\subsection{Synergy Expert}
\label{sec:synergy_expert}

In Phase II, we introduce the Synergy Expert to augment initial semantic representations by mining local topological contexts. Unlike conventional Graph Neural Networks (GNNs) that perform passive neighborhood aggregation, this expert implements an instruction-driven retrieval mechanism, utilizing semantic intent to achieve active alignment and filtering of structural information.

\subsubsection{Density-Aware Identity Anchoring.} A fundamental challenge in structural integration is the trade-off between representation collapse in sparse graphs and topological noise in dense ones. To address this, we propose a \textit{density-aware identity anchoring} strategy to construct the candidate context pool $\mathcal{C}_x$ for a given entity $x$:

\begin{equation}
    \mathcal{C}_{x} = \mathcal{A}_{self} \cup \left\{ \mathbf{e}_{n_j}^{sem} \mid n_j \in \mathcal{N}(x) \right\},
\end{equation}

where the identity anchor $\mathcal{A}_{self}$ is adaptively activated based on a node-level density threshold $\phi$:

\begin{equation}
    \mathcal{A}_{self} =
    \begin{cases}
    \left\{ \mathbf{e}_{x}^{sem} \right\}, & \text{if } |\mathcal{N}(x)| \leq \phi, \\
    \emptyset, & \text{if } |\mathcal{N}(x)| > \phi.
    \end{cases}
    \label{eq:anchor}
\end{equation}

Specifically, we set $\phi=1$ for dense graphs (e.g., FB15k-237) to prioritize extrinsic structural signals, while for sparse graphs (e.g., WN18RR), the anchoring mechanism is maintained to preserve inherent semantic stability.

\textit{Insight:} This strategy provides a semantic safety net for sparse entities to prevent manifold drift during aggregation, while forcing dense entities to explore extrinsic knowledge by decoupling redundant self-loops to mitigate over-smoothing.

\subsubsection{Cross-Modal Synergy Attention.}
To facilitate precise structural retrieval guided by semantic intent, we implement a cross-modal attention mechanism that treats semantic embeddings as query instructions. Formally, the semantic vector $\mathbf{e}_{x}^{sem}$ from Phase I is projected into a query instruction $\mathbf{q}$, while the candidate context pool $\mathcal{C}_x$ is mapped into keys $\mathbf{K}$ and values $\mathbf{V}$:

\begin{equation}
    \mathbf{q} = \mathbf{W}_Q \mathbf{e}_{x}^{sem}, \quad \mathbf{K} = \mathbf{V} = \mathbf{W}_{KV} \mathbf{H}_{\mathcal{C}x},
\end{equation}

where $\mathbf{W}_Q$ and $\mathbf{W}_{KV} \in \mathbb{R}^{d \times d}$ are learnable projection matrices, and $\mathbf{H}_{\mathcal{C}x} \in \mathbb{R}^{|\mathcal{C}_x| \times d}$ denotes the stacked embeddings of the structural neighborhood. The synergy context $\mathbf{c}_{syn}$ is derived by aligning the semantic intent with the structural background through a scaled dot-product attention:

\begin{equation}
    \mathbf{c}_{syn} = \text{softmax}\left( \frac{\mathbf{q}\mathbf{K}^\top}{\sqrt{d}} \right) \mathbf{V}.
\end{equation}

\subsubsection{Adaptive Gating and Fusion.}
Recognizing that modality reliability fluctuates across heterogeneous triplets, we introduce an MLP-based adaptive gating coefficient $\alpha$ to dynamically modulate the structural-semantic trade-off. We first compute the gated synergy vector $\mathbf{h}_{syn}$ as follows:
\begin{equation}
    \mathbf{h}_{syn} = \alpha \mathbf{q} + (1 - \alpha) \mathbf{c}_{syn},
\end{equation}
where the gating coefficient $\alpha \in [0, 1]$ is derived via a non-linear projection of the concatenated cross-modal features:
\begin{equation}
    \alpha = \sigma \left( \mathbf{w}_{g2}^\top \tanh \left( \mathbf{W}_{g1} [\mathbf{q} ; \mathbf{c}_{syn}] + \mathbf{b}_{g1} \right) + b_{g2} \right).
\end{equation}
Finally, the synergy representation $\Phi(x)$ is synthesized via a residual connection followed by layer normalization to ensure manifold stability:
\begin{equation}
    \Phi(x) = \text{LayerNorm}\left( \mathbf{e}_{x}^{sem} + \text{Dropout}\left( \mathbf{h}_{syn} \right) \right).
\end{equation}
\textit{Implementation Note:} In the gating module, $b_{g2}$ is initialized to a positive constant (e.g., $2.0$) to ensure semantic dominance during the initial stage of Phase II. This specialized initialization serves as a warm-start mechanism, preventing representation collapse when the Synergy Expert begins to integrate potentially noisy topological signals.

\begin{figure*}[!t]
  \centering
  \includegraphics[width=\linewidth]{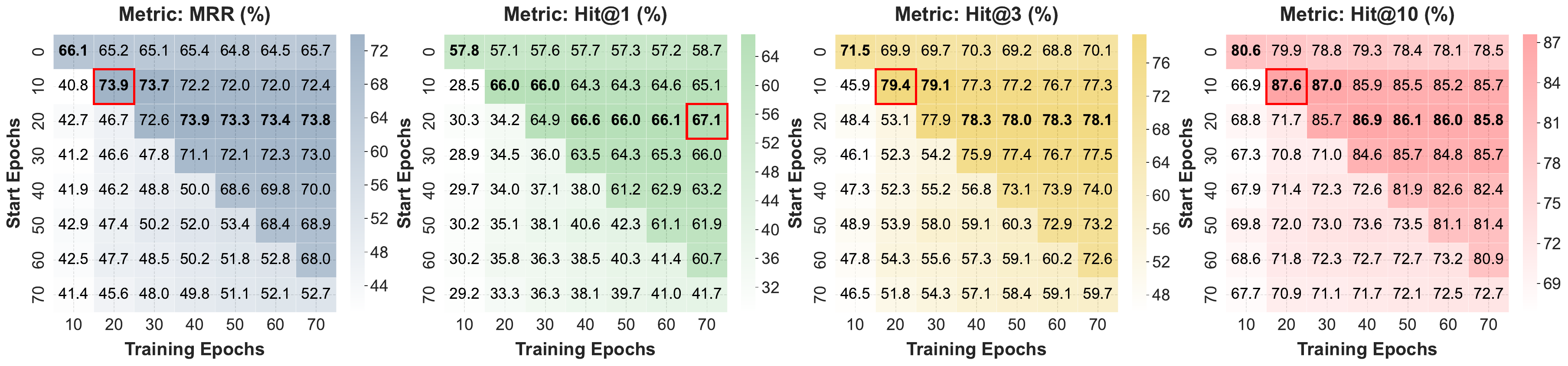}
  \caption{Analysis of synergy activation thresholds (WN18RR). The heatmaps illustrate the performance landscape across varied start epochs (vertical axis) and total training epochs (horizontal axis). Red boxes highlight the global optima across the entire grid, while bold values denote the peak performance within each column, indicating the optimal activation timing for a given training budget. Notably, activation at Epoch 20 consistently yields the most stable and superior results across all metrics, justifying its selection as the default threshold.}
  \label{fig:fusion_timing_heatmap}
\end{figure*}

\begin{table}[!t]
  \centering
  \scriptsize
  \renewcommand{\arraystretch}{1.3}
  \setlength{\tabcolsep}{4.0pt}

  \definecolor{colorBlue}{HTML}{446A91}
  \definecolor{colorGreen}{HTML}{6DC170}
  \definecolor{colorYellow}{HTML}{E6B400}
  \definecolor{colorRed}{HTML}{FF4D4F}
  \definecolor{colorGray}{HTML}{4B5563}
  \definecolor{graytxt}{HTML}{4B5563}
  \newcommand{\deep}{38}
  \newcommand{\light}{10}

  \begin{tabular}{ll ccccc}
    \toprule
    & & & \multicolumn{3}{c}{\textbf{Neighbor Hops}} & \\
    \cmidrule(lr){3-7}
    \textbf{Dataset} & \textbf{Metric} & \textbf{Hop 1} & \textbf{Hop 2} & \textbf{Hop 3} & \textbf{Hop 4} & \textbf{Hop 5} \\
    \midrule

    \multirow{2}{*}{\shortstack[l]{\textbf{FB15k-237}\\\textit{\textcolor{graytxt}{(Dense)}}}}
    & \textbf{MRR (\%)}    & \cellcolor{colorBlue!\light}37.6 & \cellcolor{colorBlue!\deep}\textbf{39.9} & \cellcolor{colorBlue!\light}39.3 & \cellcolor{colorBlue!\light}39.3 & \cellcolor{colorBlue!\light}39.3 \\
    & \textbf{Hits@1 (\%)} & \cellcolor{colorGreen!\light}27.7 & \cellcolor{colorGreen!\deep}\textbf{30.2} & \cellcolor{colorGreen!\light}29.7 & \cellcolor{colorGreen!\light}29.7 & \cellcolor{colorGreen!\light}29.6 \\
    & \textbf{Hits@3 (\%)} & \cellcolor{colorYellow!\light}41.3 & \cellcolor{colorYellow!\deep}\textbf{43.6} & \cellcolor{colorYellow!\light}42.8 & \cellcolor{colorYellow!\light}42.8 & \cellcolor{colorYellow!\light}42.8 \\
    & \textbf{Hits@10 (\%)}& \cellcolor{colorRed!\light}57.6 & \cellcolor{colorRed!\deep}\textbf{59.4} & \cellcolor{colorRed!\light}58.5 & \cellcolor{colorRed!\light}58.5 & \cellcolor{colorRed!\light}58.5 \\
    & \textbf{Mean Rank}      & \cellcolor{colorGray!\light}85.7 & \cellcolor{colorGray!\deep}\textbf{81.4} & \cellcolor{colorGray!\light}84.2 & \cellcolor{colorGray!\light}84.3 & \cellcolor{colorGray!\light}84.3 \\

    \midrule

    \multirow{2}{*}{\shortstack[l]{\textbf{WN18RR}\\\textit{\textcolor{graytxt}{(Sparse)}}}}
    & \textbf{MRR (\%)}    & \cellcolor{colorBlue!\deep}\textbf{74.2} & \cellcolor{colorBlue!\light}73.5 & \cellcolor{colorBlue!\light}73.7 & \cellcolor{colorBlue!\light}73.6 & \cellcolor{colorBlue!\light}73.7 \\
    & \textbf{Hits@1 (\%)} & \cellcolor{colorGreen!\deep}\textbf{67.7} & \cellcolor{colorGreen!\light}66.8 & \cellcolor{colorGreen!\light}66.8 & \cellcolor{colorGreen!\light}66.7 & \cellcolor{colorGreen!\light}67.0 \\
    & \textbf{Hits@3 (\%)} & \cellcolor{colorYellow!\deep}\textbf{78.5} & \cellcolor{colorYellow!\light}77.9 & \cellcolor{colorYellow!\light}78.1 & \cellcolor{colorYellow!\light}78.1 & \cellcolor{colorYellow!\light}77.9 \\
    & \textbf{Hits@10 (\%)}& \cellcolor{colorRed!\light}85.5 & \cellcolor{colorRed!\light}85.5 & \cellcolor{colorRed!\deep}\textbf{86.0} & \cellcolor{colorRed!\light}85.8 & \cellcolor{colorRed!\light}85.7 \\
    & \textbf{Mean Rank}      & \cellcolor{colorGray!\light}109.8 & \cellcolor{colorGray!\light}106.5 & \cellcolor{colorGray!\light}104.2 & \cellcolor{colorGray!\light}101.4 & \cellcolor{colorGray!\deep}\textbf{93.2} \\

    \bottomrule
  \end{tabular}
  \vspace{6pt}
  \caption{Topological scaling analysis across neighbor hop depths (1--5). Cell shading denotes metric palettes; deep shading and bold values identify optima. The results highlight a density-driven trade-off: dense FB15k-237 favors a 2-hop local receptive field to resolve relational ambiguity, whereas sparse WN18RR exhibits Top-k saturation at Hop 1 but achieves significant global ordering refinement (Mean Rank) via deep 5-hop topological signals.}
  \label{tab:hops_unified_style_single_column}
\end{table}

\begin{figure*}[!t]
  \centering
  \begin{subfigure}{0.613\textwidth}
    \centering
    \includegraphics[width=\linewidth]{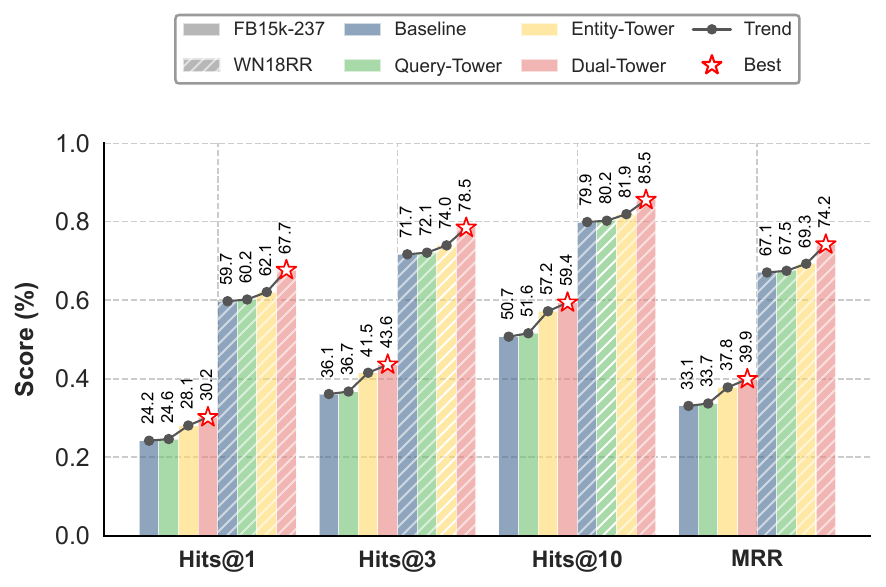}
    \caption{Intra-tower Consistency.}
    \label{fig:ablation_architectural}
  \end{subfigure}
  \hfill
  \begin{subfigure}{0.37\textwidth}
    \centering
    \includegraphics[width=\linewidth]{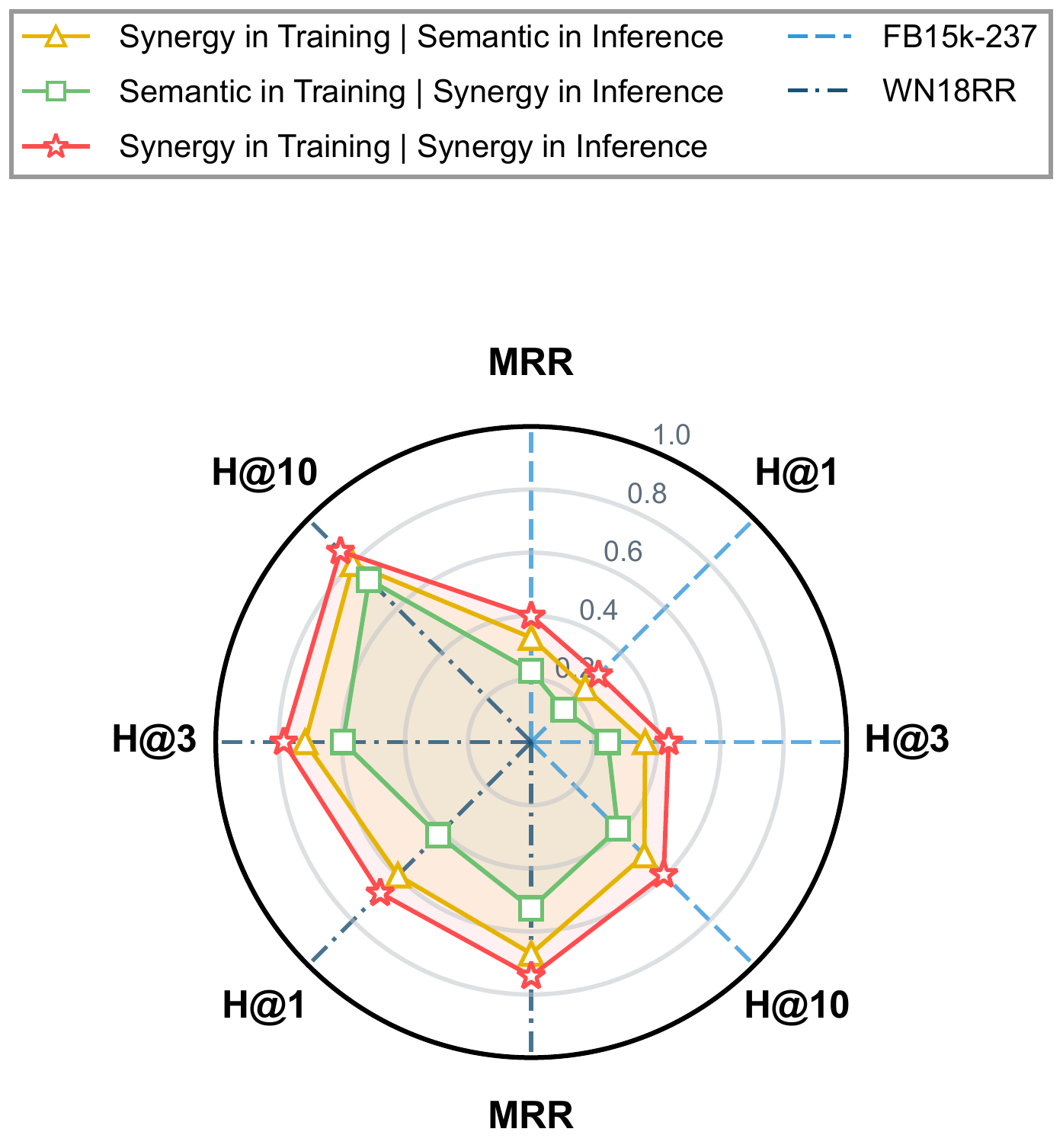}
    \caption{Cross-phase Consistency.}
    \label{fig:ablation_lifecycle}
  \end{subfigure}

  \caption{Ablation analysis of the proposed Dual-Tower Synergy Mechanism. Left: Evaluation of Architectural Synergy reveals that the structural consistency between the Query and Entity towers is the prerequisite for aligning head-tail semantic manifolds. Right: Investigation into Lifecycle Synergy demonstrates that keeping the synergy mechanism consistent across training and inference stages is critical for mitigating the distribution shift, thereby ensuring that learned collaborative features are fully activated during deployment.}
  \label{fig:synergy_ablation_main}
\end{figure*}

\begin{table}[!t]
  \centering
  \scriptsize
  \definecolor{cAlign}{HTML}{446A91}
  \definecolor{cCross}{HTML}{6DC170}
  \definecolor{cGate}{HTML}{E6B400}
  \definecolor{cSynergy}{HTML}{FF4D4F}
  \definecolor{incrc}{HTML}{B91C1C}
  \definecolor{graytxt}{HTML}{4B5563}
  \definecolor{decrc}{HTML}{6DC170}

  \definecolor{colorBlue}{HTML}{446A91}
  \definecolor{colorGreen}{HTML}{6DC170}
  \definecolor{colorYellow}{HTML}{E6B400}
  \definecolor{colorRed}{HTML}{FF4D4F}
  \definecolor{colorGray}{HTML}{4B5563}
  \newcommand{\deep}{38}
  \newcommand{\light}{10}

  \renewcommand{\arraystretch}{1.3}
  \setlength{\tabcolsep}{2.2pt}

  \begin{tabular}{llcccc}
    \toprule
    & & \multicolumn{3}{c}{\textbf{Ablated Variants}} & \\
    \cmidrule(lr){3-5}
    \textbf{Dataset} & \textbf{Metric (\%)} & \textbf{w/o Align} & \textbf{w/o Cross} & \textbf{w/o Gate} & \textbf{SynergyKGC} \\
    \midrule

    \multirow{2}{*}{\shortstack[l]{\textbf{FB15k-237}\\\textit{\textcolor{graytxt}{(Hop 2)}}}}
    & \textbf{MRR}     & \cellcolor{colorBlue!\light}38.2 \textcolor{decrc}{\textit{$\downarrow$1.7}} & \cellcolor{colorBlue!\light}38.0 \textcolor{decrc}{\textit{$\downarrow$1.9}} & \cellcolor{colorBlue!\light}36.8 \textcolor{decrc}{\textit{$\downarrow$3.1}} & \cellcolor{colorBlue!\deep}\textbf{39.9} \\
    & \textbf{Hits@1}  & \cellcolor{colorGreen!\light}28.5 \textcolor{decrc}{\textit{$\downarrow$1.7}} & \cellcolor{colorGreen!\light}28.4 \textcolor{decrc}{\textit{$\downarrow$1.8}} & \cellcolor{colorGreen!\light}27.3 \textcolor{decrc}{\textit{$\downarrow$2.9}} & \cellcolor{colorGreen!\deep}\textbf{30.2} \\
    & \textbf{Hits@3}  & \cellcolor{colorYellow!\light}41.9 \textcolor{decrc}{\textit{$\downarrow$1.7}} & \cellcolor{colorYellow!\light}41.5 \textcolor{decrc}{\textit{$\downarrow$2.1}} & \cellcolor{colorYellow!\light}40.3 \textcolor{decrc}{\textit{$\downarrow$3.3}} & \cellcolor{colorYellow!\deep}\textbf{43.6} \\
    & \textbf{Hits@10} & \cellcolor{colorRed!\light}57.6 \textcolor{decrc}{\textit{$\downarrow$1.8}} & \cellcolor{colorRed!\light}57.3 \textcolor{decrc}{\textit{$\downarrow$2.1}} & \cellcolor{colorRed!\light}55.7 \textcolor{decrc}{\textit{$\downarrow$3.7}} & \cellcolor{colorRed!\deep}\textbf{59.4} \\

    \midrule

    \multirow{2}{*}{\shortstack[l]{\textbf{WN18RR}\\\textit{\textcolor{graytxt}{(Hop 1)}}}}
    & \textbf{MRR}     & \cellcolor{colorBlue!\light}72.6 \textcolor{decrc}{\textit{$\downarrow$1.6}} & \cellcolor{colorBlue!\light}52.9 \textcolor{decrc}{\textit{$\downarrow$21.3}} & \cellcolor{colorBlue!\light}60.5 \textcolor{decrc}{\textit{$\downarrow$13.7}} & \cellcolor{colorBlue!\deep}\textbf{74.2} \\
    & \textbf{Hits@1}  & \cellcolor{colorGreen!\light}66.0 \textcolor{decrc}{\textit{$\downarrow$1.7}} & \cellcolor{colorGreen!\light}42.0 \textcolor{decrc}{\textit{$\downarrow$25.7}} & \cellcolor{colorGreen!\light}51.2 \textcolor{decrc}{\textit{$\downarrow$16.5}} & \cellcolor{colorGreen!\deep}\textbf{67.7} \\
    & \textbf{Hits@3}  & \cellcolor{colorYellow!\light}76.6 \textcolor{decrc}{\textit{$\downarrow$1.9}} & \cellcolor{colorYellow!\light}59.4 \textcolor{decrc}{\textit{$\downarrow$19.1}} & \cellcolor{colorYellow!\light}66.3 \textcolor{decrc}{\textit{$\downarrow$12.2}} & \cellcolor{colorYellow!\deep}\textbf{78.5} \\
    & \textbf{Hits@10} & \cellcolor{colorRed!\light}84.5 \textcolor{decrc}{\textit{$\downarrow$1.0}} & \cellcolor{colorRed!\light}73.2 \textcolor{decrc}{\textit{$\downarrow$12.3}} & \cellcolor{colorRed!\light}77.5 \textcolor{decrc}{\textit{$\downarrow$8.0}} & \cellcolor{colorRed!\deep}\textbf{85.5} \\

    \bottomrule
  \end{tabular}

  \vspace{6pt}
  \caption{Ablation analysis of \textit{SynergyKGC} components at optimal topological depths. Bold values denote peak performance; green markers with $\downarrow$ quantify the performance degradation relative to the full model. The disparate sensitivity across datasets underscores the necessity of each module for robust inference across varying graph densities.}
  \label{tab:synergy_transposed_ablation}
\end{table}

\subsection{Dynamic Dual-Tower Consistency Inference}
To mitigate the representation shift between training and evaluation phases, \textit{SynergyKGC} employs a Dynamic Dual-Tower Consistency mechanism. Unlike conventional methods that rely on static, decoupled embeddings for candidate entities during testing, our approach enforces the simultaneous activation of the Synergy Expert for both the Query Tower and the Entity Tower to maintain manifold alignment.

For a given query $(h, r)$ and any candidate tail entity $t \in \mathcal{E}$, the final representations are generated in real-time via the synergy function $\Phi$ defined in Section~\ref{sec:synergy_expert}:
\begin{equation}
    \mathbf{e}_{hr}^{syn} = \Phi_{hr}(h, r), \quad \mathbf{e}_{t}^{syn} = \Phi_{t}(t), \quad \forall t \in \mathcal{E}.
\end{equation}
\textit{Architectural Parity:} Crucially, \textit{SynergyKGC} maintains strict dual-tower synergy consistency across both training and inference phases. Once the Synergy Expert is activated, the identical structural-semantic reconciliation logic is applied to both learning and prediction, ensuring that the scoring function $\psi$ operates within a unified, synergy-enhanced manifold:
\begin{equation}
    \psi(h, r, t) = \frac{(\mathbf{e}_{hr}^{syn})^\top \mathbf{e}_{t}^{syn}}{\| \mathbf{e}_{hr}^{syn} \|_2 \| \mathbf{e}_{t}^{syn} \|_2}.
\end{equation}

This fully dynamic design ensures that candidates capture their latest topological contexts, achieving precise alignment within a consistent high-dimensional space.

\subsection{Joint Optimization Objective}
In the second stage of our strategy (Phase II), \textit{SynergyKGC} is optimized via a multi-task objective to reconcile high-precision retrieval with manifold stability.

\noindent \textbf{Synergy Contrastive Loss.} To maximize the compatibility of positive triplets in the synergy-enhanced space, we apply a contrastive warming objective over batch $\mathcal{B}$~\cite{oord:arxiv18-cpc}:
\begin{equation}
    \mathcal{L}_{NCE}^{syn} = - \sum_{(h,r,t) \in \mathcal{B}} \log \frac{e^{s_{syn}(h,r,t) / \tau}}{\sum_{t' \in \mathcal{B}_{neg} \cup \{t\}} e^{s_{syn}(h,r,t') / \tau}},
\end{equation}
where $s_{syn}$ represents the scoring function in the collaborative manifold.

\noindent \textbf{Semantic Consistency Regularization.} To prevent the semantic drift of the pre-established feature manifold during structural injection, we introduce an alignment loss based on Mean Squared Error (MSE). This constraint ensures that the synergy-enhanced representations $\Phi(e)$ remain proximal to their semantic anchors $\mathbf{e}_{e}^{sem}$ solidified in Phase I:
\begin{equation}
    \mathcal{L}_{align}^{x} = \frac{1}{|\mathcal{B}|} \sum_{e \in \mathcal{B}_x} \| \Phi(e) - \text{sg}(\mathbf{e}_{e}^{sem}) \|_2^2, \quad x \in \{hr, t\},
\end{equation}
where $\text{sg}(\cdot)$ denotes the stop-gradient operation (implemented via cloning and detaching) to preserve the base semantic expert.

\noindent \textbf{Total Objective.} The joint loss for Phase II is activated when $\text{epoch} \geq T_{start}$ (where $T_{start}$ is 5 for FB15k-237 and 20 for WN18RR), formulated as:
\begin{equation}
    \mathcal{L}_{total} = \mathcal{L}_{NCE}^{syn} + \lambda \cdot (\mathcal{L}_{align}^{hr} + \mathcal{L}_{align}^{t}),
\end{equation}
where $\lambda = 0.1$ is the penalty weight that balances retrieval precision and manifold stability.

\section{Experiments}

\subsection{Experimental Setup}

\noindent \textbf{Datasets.} We conduct experiments on two benchmark datasets, FB15k-237~\cite{toutanova:cvsc15-observed,bollacker:sigmod08-freebase} and WN18RR\cite{dettmers:aaai18-conve,miller:cacm95-wordnet}, which offer polarized structural and semantic landscapes to rigorously test the robustness of our \textit{SynergyKGC} (Table~\ref{tab:dataset_stats_compact}). FB15k-237 represents a relational-dense environment with high topological overlap ($P_{50}=22$) and verbose textual descriptions (avg. 114.6 tokens), requiring the model to effectively disambiguate complex relational fingerprints. In contrast, WN18RR provides a hierarchical-sparse testbed characterized by minimal connectivity ($P_{50}=3$) and succinct semantics (avg. 17.1 tokens), emphasizing the model's capacity to anchor sparse relational cues. This divergent selection allows us to evaluate \textit{SynergyKGC}'s effectiveness in adaptively reconciling semantic manifolds with explicit topological signals across heterogeneous graph densities.

\noindent \textbf{Evaluation Metrics.} We evaluate performance using the standard filtered setting across five ranking-based metrics: Mean Reciprocal Rank (MRR), Mean Rank (MR), and Hits@$k$ ($k \in \{1, 3, 10\}$). While MRR and Hits@$k$ prioritize the model's precision in retrieving top-tier candidates, the inclusion of MR serves as a critical measure of global ranking stability. This comprehensive suite allows us to verify \textit{SynergyKGC}'s dual capacity: (i) achieving high-precision alignment between semantic and structural signals (reflected in MRR/Hits), and (ii) effectively constraining the overall search space to resolve long-tail ambiguity across heterogeneous graph densities (reflected in MR).

\noindent \textbf{Implementation Details.} \textit{SynergyKGC} is implemented using \textit{bert-base-uncased} and a 4-head synergy expert ($d=768$) as semantic and structural encoders, respectively. Training is conducted on a single NVIDIA A100 GPU with a batch size of 768, using the AdamW optimizer with a learning rate selected from $\{1e\text{-}5, 5e\text{-}5, 5e\text{-}4\}$. We set $\tau=0.05$, $\gamma=0.02$, and apply a 0.1 dropout rate globally. A two-phase training strategy is adopted, activating the synergy expert at Epoch 5 for FB15k-237 and Epoch 20 for WN18RR. To stabilize modality transition, an alignment auxiliary loss (weight 0.1) is introduced in Phase 2. Furthermore, to adaptively manage topological heterogeneity, the density threshold $\phi$ is set to 1 for FB15k-237, whereas no threshold is applied to WN18RR to preserve its inherent hierarchical structure. Further implementation details are available in our released code.

\subsection{Main Results}

As detailed in Table~\ref{tab:main_results}, \textit{SynergyKGC} establishes a new state-of-the-art across both dense (FB15k-237) and sparse (WN18RR) datasets, delivering significant improvements over top-tier hybrid models such as \textit{ProgKGC}~\cite{li:iswc25-progkgc}. Notably, the absolute gain of $+8.0\%$ in Hits@1 on WN18RR underscores the model's exceptional precision in low-density topological environments. This performance leap is fundamentally attributed to the bidirectional synchronization mechanism; as visualized in Fig.~\ref{fig:mrr_comparison}, the activation of the \textit{Synergy Expert} triggers an instantaneous ``catch-up'' effect, abruptly eliminating the representation divergence between the forward and backward streams. By reconciling disparate modal signals into a coherent, high-fidelity manifold, \textit{SynergyKGC} effectively mitigates the distribution shift typically encountered in late-phase training, thereby ensuring robust and precise link prediction.

\subsection{Ablation Study}

\begin{table*}[!t]
  \centering
  \scriptsize

  \definecolor{colorBlue}{HTML}{446A91}
  \definecolor{colorGreen}{HTML}{6DC170}
  \definecolor{colorYellow}{HTML}{E6B400}
  \definecolor{colorRed}{HTML}{FF4D4F}
  \definecolor{colorGray}{HTML}{4B5563}
  \newcommand{\deep}{38}
  \newcommand{\light}{10}

  \setlength{\tabcolsep}{8pt}
  \begin{tabular}{@{} c c @{}}

    \begin{minipage}[t]{0.70\textwidth}
      \vspace{0pt}
      \centering
      \renewcommand{\arraystretch}{1.2}

      \setlength{\tabcolsep}{3.4pt}

      \begin{tabular}{llcccccccccccc}
        \toprule
        & & & \multicolumn{11}{c}{\textbf{Density Threshold ($T$)}} \\
        \cmidrule(lr){4-14}
        \textbf{Dataset} & \textbf{Metric(\%)} & \textbf{w/o Anchor} & \textbf{1} & \textbf{2} & \textbf{3} & \textbf{4} & \textbf{5} & \textbf{8} & \textbf{11} & \textbf{22} & \textbf{41} & \textbf{482} & \textbf{7614} \\
        \midrule
        & \textcolor{colorGray}{\textit{Percentile}} & & \textcolor{colorGray}{$P_0$} & & & & \textcolor{colorGray}{$P_{10}$} & & \textcolor{colorGray}{$P_{25}$} & \textcolor{colorGray}{$P_{50}$} & \textcolor{colorGray}{$P_{75}$} & & \textcolor{colorGray}{$P_{100}$} \\
        \multirow{2}{*}{\shortstack[l]{\textbf{FB15k-237}\\\textit{\textcolor{colorGray}{(Hop 2)}}}}
        & \textbf{MRR}    & \cellcolor{colorBlue!\light}39.3 & \cellcolor{colorBlue!\deep}\textbf{39.9} & \cellcolor{colorBlue!\light}39.4 & \cellcolor{colorBlue!\light}38.8 & \cellcolor{colorBlue!\light}38.0 & \cellcolor{colorBlue!\light}37.7 & \cellcolor{colorBlue!\light}37.8 & \cellcolor{colorBlue!\light}37.2 & \cellcolor{colorBlue!\light}37.6 & \cellcolor{colorBlue!\light}38.2 & \cellcolor{colorBlue!\light}--- & \cellcolor{colorBlue!\light}38.4 \\
        & \textbf{Hits@1}  & \cellcolor{colorGreen!\light}29.6 & \cellcolor{colorGreen!\deep}\textbf{30.2} & \cellcolor{colorGreen!\light}29.6 & \cellcolor{colorGreen!\light}29.1 & \cellcolor{colorGreen!\light}28.2 & \cellcolor{colorGreen!\light}27.9 & \cellcolor{colorGreen!\light}28.2 & \cellcolor{colorGreen!\light}27.8 & \cellcolor{colorGreen!\light}27.8 & \cellcolor{colorGreen!\light}28.7 & \cellcolor{colorGreen!\light}--- & \cellcolor{colorGreen!\light}28.8 \\
        & \textbf{Hits@3}  & \cellcolor{colorYellow!\light}43.1 & \cellcolor{colorYellow!\deep}\textbf{43.6} & \cellcolor{colorYellow!\light}43.2 & \cellcolor{colorYellow!\light}42.5 & \cellcolor{colorYellow!\light}41.7 & \cellcolor{colorYellow!\light}41.2 & \cellcolor{colorYellow!\light}41.4 & \cellcolor{colorYellow!\light}40.7 & \cellcolor{colorYellow!\light}41.2 & \cellcolor{colorYellow!\light}41.8 & \cellcolor{colorYellow!\light}--- & \cellcolor{colorYellow!\light}41.8 \\
        & \textbf{Hits@10} & \cellcolor{colorRed!\light}58.7 & \cellcolor{colorRed!\deep}\textbf{59.4} & \cellcolor{colorRed!\light}59.0 & \cellcolor{colorRed!\light}58.4 & \cellcolor{colorRed!\light}57.6 & \cellcolor{colorRed!\light}57.1 & \cellcolor{colorRed!\light}57.1 & \cellcolor{colorRed!\light}56.3 & \cellcolor{colorRed!\light}57.3 & \cellcolor{colorRed!\light}57.6 & \cellcolor{colorRed!\light}--- & \cellcolor{colorRed!\light}57.6 \\
        \midrule
        & \textcolor{colorGray}{\textit{Percentile}} & & \textcolor{colorGray}{$P_{0}/P_{10}$} & \textcolor{colorGray}{$P_{25}$} & \textcolor{colorGray}{$P_{50}$} & & \textcolor{colorGray}{$P_{75}$} & \textcolor{colorGray}{$P_{90}$} & & & & \textcolor{colorGray}{$P_{100}$} \\
        \multirow{2}{*}{\shortstack[l]{\textbf{WN18RR}\\\textit{\textcolor{colorGray}{(Hop 1)}}}}
        & \textbf{MRR}    & \cellcolor{colorBlue!\light}50.2 & \cellcolor{colorBlue!\light}70.5 & \cellcolor{colorBlue!\light}70.6 & \cellcolor{colorBlue!\light}71.0 & \cellcolor{colorBlue!\light}71.1 & \cellcolor{colorBlue!\light}52.0 & \cellcolor{colorBlue!\light}71.0 & \cellcolor{colorBlue!\light}--- & \cellcolor{colorBlue!\light}--- & \cellcolor{colorBlue!\light}--- & \cellcolor{colorBlue!\deep}\textbf{74.2} & \cellcolor{colorBlue!\light}--- \\
        & \textbf{Hits@1}  & \cellcolor{colorGreen!\light}38.8 & \cellcolor{colorGreen!\light}63.2 & \cellcolor{colorGreen!\light}63.5 & \cellcolor{colorGreen!\light}63.7 & \cellcolor{colorGreen!\light}64.0 & \cellcolor{colorGreen!\light}40.5 & \cellcolor{colorGreen!\light}63.5 & \cellcolor{colorGreen!\light}--- & \cellcolor{colorGreen!\light}--- & \cellcolor{colorGreen!\light}--- & \cellcolor{colorGreen!\deep}\textbf{67.7} & \cellcolor{colorGreen!\light}--- \\
        & \textbf{Hits@3}  & \cellcolor{colorYellow!\light}56.4 & \cellcolor{colorYellow!\light}75.2 & \cellcolor{colorYellow!\light}75.1 & \cellcolor{colorYellow!\light}75.8 & \cellcolor{colorYellow!\light}75.5 & \cellcolor{colorYellow!\light}58.9 & \cellcolor{colorYellow!\light}75.9 & \cellcolor{colorYellow!\light}--- & \cellcolor{colorYellow!\light}--- & \cellcolor{colorYellow!\light}--- & \cellcolor{colorYellow!\deep}\textbf{78.5} & \cellcolor{colorYellow!\light}--- \\
        & \textbf{Hits@10} & \cellcolor{colorRed!\light}71.9 & \cellcolor{colorRed!\light}83.6 & \cellcolor{colorRed!\light}83.3 & \cellcolor{colorRed!\light}84.1 & \cellcolor{colorRed!\light}84.1 & \cellcolor{colorRed!\light}73.6 & \cellcolor{colorRed!\light}84.4 & \cellcolor{colorRed!\light}--- & \cellcolor{colorRed!\light}--- & \cellcolor{colorRed!\light}--- & \cellcolor{colorRed!\deep}\textbf{85.5} & \cellcolor{colorRed!\light}--- \\
        \bottomrule
      \end{tabular}
    \end{minipage}
    &
    \begin{minipage}[t]{0.25\textwidth}
      \vspace{0pt}
      \centering
      \includegraphics[width=\linewidth]{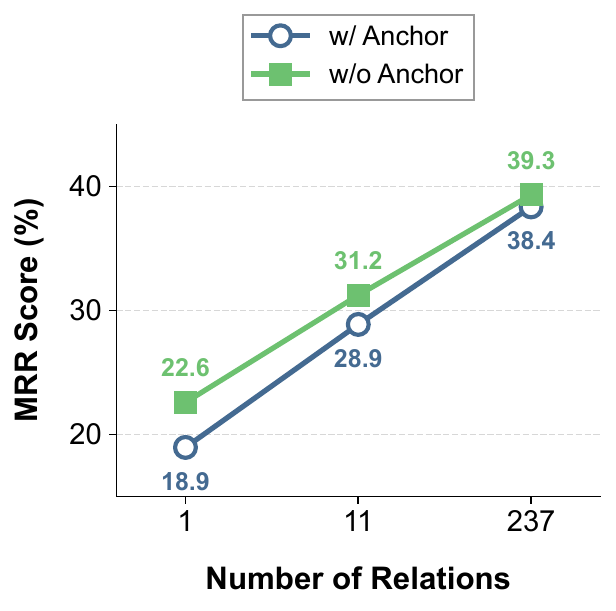}
    \end{minipage}
    \\

    \begin{minipage}[b]{0.70\textwidth}
        \vspace{-33pt}
        \captionof{table}{Sensitivity analysis of performance across varying entity density thresholds $T$, where $P_x$ denotes the $x$-th percentile of the entity degree distribution (i.e., $x\%$ of entities possess a degree $\le T$). Best results are highlighted in bold with darker color-coded backgrounds. The disparate peak performance distributions across datasets suggest that the optimal threshold is highly dependent on the underlying graph topology and sparsity.}
        \label{tab:topological_sensitivity}
    \end{minipage}
    &
    \begin{minipage}[b]{0.25\textwidth}
        \vspace{-33pt}
        \captionof{figure}{MRR performance on FB15k-237 under relational compression. The widening gap between the two configurations, even when reduced to a singular relation, demonstrates that topological density, rather than relational complexity, is the primary determinant of identity anchoring's efficacy.}
        \label{fig:relational_compression}
    \end{minipage}

  \end{tabular}
\end{table*}

\noindent \textbf{Synergy Dynamics and Topological Scaling Analysis.} The ablation studies on synergy activation and neighbor hop depth underscore the dual advantages of \textit{SynergyKGC} in convergence efficiency and structural adaptability. By significantly reducing semantic warm-up overhead, the model reaches a high-performance plateau as early as Epoch 10, contrasting sharply with the $\ge 30$ epoch requirement of paradigms like \textit{ProgKGC}, and establishes a stable global optimum at Epoch 20. This temporal efficiency is complemented by a density-driven topological scaling: while the relational-dense FB15k-237 achieves optimal disambiguation within a 2-hop local receptive field, the sparse WN18RR exhibits a unique decoupling where Hop 1 anchors top-tier precision (MRR), but deep 5-hop signals are indispensable for refining global ordering (Mean Rank). Collectively, these findings confirm the model's superior capacity to reconcile latent semantic manifolds with explicit multi-hop structural constraints, ensuring rapid convergence without the risk of representation drift. Consequently, we activate the Synergy Expert at Epoch 20 and adopt Hop 2 for FB15k-237 and Hop 1 for WN18RR as our default experimental configurations.

\noindent \textbf{Analysis of Dual-Tower Coherent Consistency.} The ablation analysis of our proposed Dual-Tower Synergy Mechanism reveals a dual-axis imperative for achieving coherent representation alignment in Knowledge Graph Completion (KGC). On the architectural axis (Figure \ref{fig:ablation_architectural}), the results underscore that structural consistency between the query and entity towers is not merely an optimization but a prerequisite for the precise alignment of head-tail semantic manifolds, where the ``Full Synergy'' configuration markedly outperforms asymmetric variants. Parallel to this, our investigation into lifecycle synergy (Figure \ref{fig:ablation_lifecycle}) identifies a critical ``cross-phase consistency'' requirement; specifically, maintaining the synergy mechanism uniformly across training and inference stages is indispensable for mitigating distribution shifts, ensuring that collaborative features meticulously learned during training are fully activated during deployment. Collectively, these findings validate the architectural and temporal robustness of \textit{SynergyKGC}, confirming that holistic synergy—spanning both model topology and the training-inference lifecycle—is the fundamental driver for reconciling relational specificity with global semantic stability.

\noindent \textbf{Ablation Analysis of Key Synergy Expert Modules.} The ablation of core functional modules within the Synergy Expert, namely MSE Alignment Loss (Align), Cross-Attention (Cross), and Adaptive Gating (Gate), reveals their differentiated roles in reconciling structural heterogeneity (Table \ref{tab:synergy_transposed_ablation}). While the Align module provides the foundational semantic consistency for our Double-tower architecture, the Cross and Gate modules function as adaptive regulators of topological information. On the sparse WN18RR dataset, the catastrophic collapse in performance following the removal of the Cross module—where MRR and Hits@1 plunge by $21.3\%$ and $25.7\%$ respectively—identifies cross-modal interaction as an essential ``positional scaffold'' for anchoring abstract relational cues. Conversely, the dense FB15k-237 dataset exhibits a heightened sensitivity to the Gate module (MRR drop of $2.9\%$), validating its role as a critical filter that suppresses identity-heavy structural noise in high-degree clusters to prevent representation collapse. This disparate impact across datasets mathematically justifies the necessity of our topology-aware architecture, proving that \textit{SynergyKGC} effectively modulates the trade-off between semantic specificity and topological stability through its modular coordination.

\begin{figure}[t]
  \centering
  \includegraphics[width=\linewidth]{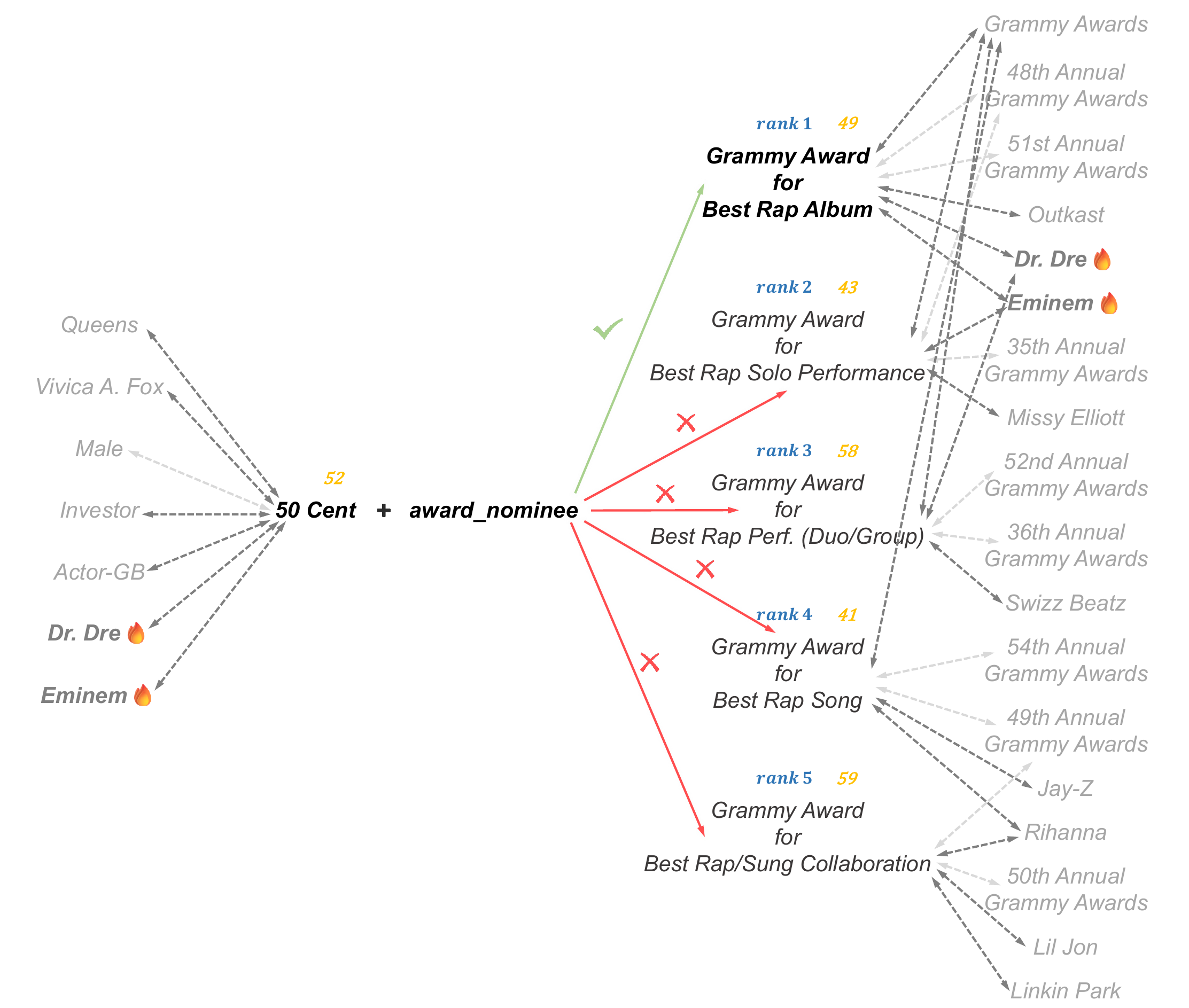}
  \caption{Structural sufficiency in FB15k-237. For \textit{50 Cent} + \textit{award\_nominee}, dual bridge entities (\textit{Dr. Dre}, \textit{Eminem}) form a closed relational path between high-degree nodes, enabling precise localization independent of identity anchoring and validating the ``Structure $\approx$ Identity'' phenomenon.}
  \label{fig:example1}
\end{figure}

\begin{figure}[t]
  \centering
  \includegraphics[width=\linewidth]{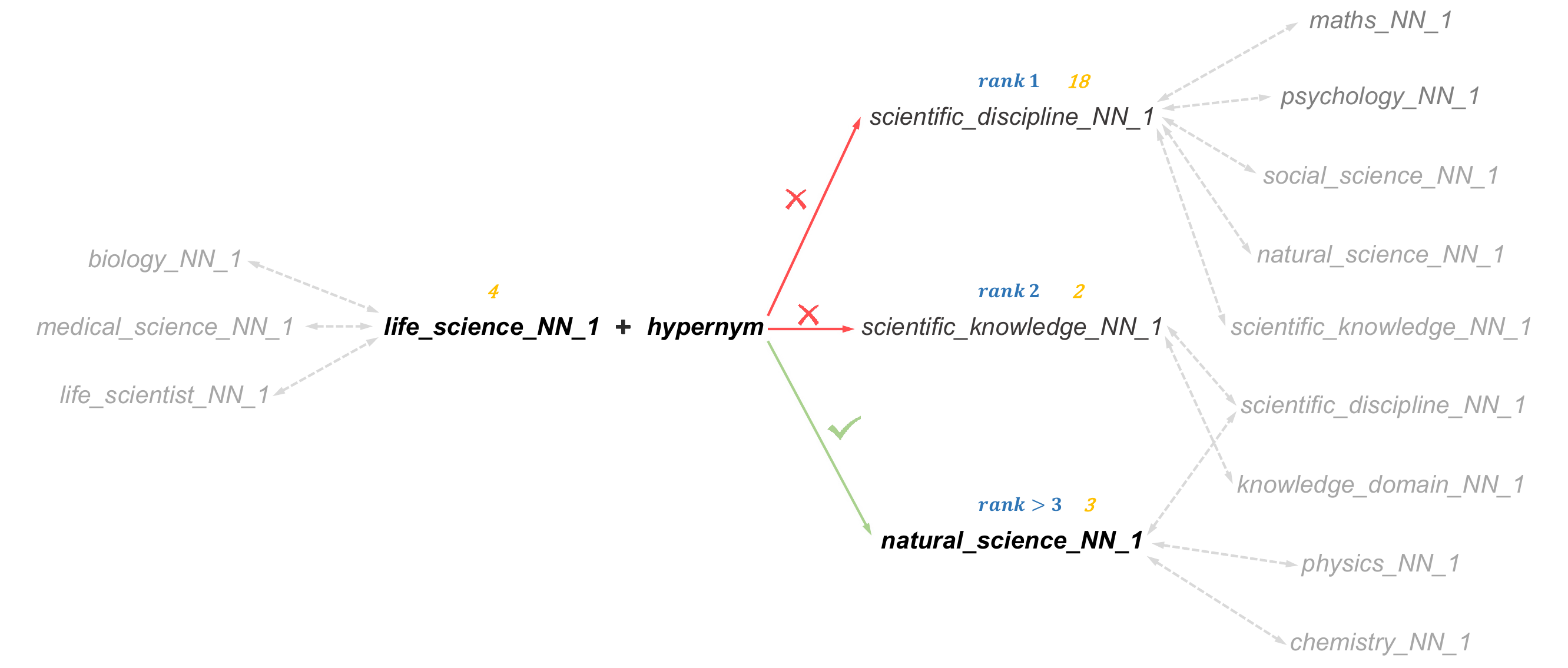}
  \caption{Popularity bias and structural scaffolding in WN18RR. For the query \textit{life\_science} + \textit{hypernym}, the sparse target (degree 3) and the hub-like distractor (\textit{maths}, degree 73) are topologically indistinguishable via a shared generic bridge. The Identity Anchoring strategy serves as a critical scaffold to overcome this ``degree trap.''}
  \label{fig:example2}
\end{figure}

\noindent \textbf{Mechanisms of Identity Anchoring.} The analysis of our proposed Identity Anchoring (IA) strategy reveals a fundamental structural resolution mismatch across heterogeneous graph topologies, as evidenced by the divergent performance sensitivity in Table \ref{tab:topological_sensitivity}, where IA acts as essential positional scaffolding in sparse environments like WN18RR, achieving a peak MRR of 74.2\%, yet introduces irreducible structural noise in dense benchmarks like FB15k-237, where optimal results are restricted to minimal thresholds. To further isolate the underlying drivers of this phenomenon, our relational compression experiments in Figure \ref{fig:relational_compression} demonstrate a state of topological determinism; by stripping relational diversity from 237 down to a singular relation, we observe that the performance gap between anchored and ablated configurations persists and even widens, revealing that relational richness previously served as a semantic buffer masking the detrimental effects of identity-heavy cues. This evidence demonstrates that the efficacy of the anchoring strategy is governed by intrinsic topological features rather than relational schema complexity alone, thereby justifying the necessity of the topology-aware synergy mechanism within \textit{SynergyKGC}.

\noindent \textbf{Qualitative Analysis of Representative Case Studies.}
In the FB15k-237 (dense) case (Figure \ref{fig:example1}), within our Double-tower Coherent Consistency architecture, the Cross-Modal Synergy Expert identifies an ``over-complete'' structural fingerprint, comprising specific neighbors like \textit{Queens}, \textit{Male}, and \textit{Investor}, that uniquely localizes the head entity (\textit{50 Cent}, degree 52) without relying on explicit IDs. Log analysis confirms that dual bridge entities (\textit{Dr. Dre}, \textit{Eminem}) form a closed reasoning path to the target (degree 49), further validating the Structure $\approx$ Identity phenomenon . Conversely, the \textit{life\_science} + \textit{hypernym} query in the sparse WN18RR hierarchy (Figure \ref{fig:example2}) highlights a critical Popularity Bias. Here, the Cross-Modal Synergy Expert detects the entity's extreme sparsity (degree 4) and its reliance on generic bridges. Without IA, the model falls into a ``degree trap'' due to topological isomorphism, favoring hub-like distractors (\textit{maths}, degree 73) over the sparse target (\textit{natural\_science}, degree 3) . In such long-tailed distributions, IA serves as essential positional scaffolding to prevent representation drift. Ultimately, these prove that anchoring efficacy is governed by intrinsic topological density rather than relational schema complexity alone, justifying our topology-aware synergy mechanism.

\section{Conclusion}

\noindent In this paper, we presented \textit{SynergyKGC}, a novel dual-tower framework that effectively reconciles latent semantic manifolds with explicit topological signals through a cross-modal Synergy Expert. By implementing a two-phase training strategy with density-aware identity anchoring—utilizing $\phi=1$ for dense graphs and preserved connectivity for sparse ones—we successfully eliminated the representation gap between dual-tower streams while significantly reducing semantic warm-up overhead. Our extensive evaluations on FB15k-237 and WN18RR, highlighted by a +8.0\% gain in Hits@1, demonstrate the model's superior robustness and efficiency across heterogeneous graph densities. This work underscores the critical role of adaptive structural-semantic reconciliation in preventing representation drift and provides a scalable paradigm for high-precision Knowledge Graph Completion.

To facilitate reproducibility, we have made the source code of SynergyKGC, including all model architectures, training scripts, and pre-processed datasets, publicly available at: https://github.com/XuechengZou-2001/SynergyKGC-main.

\bibliographystyle{kr}
\bibliography{kr-sample}

\end{document}